\pdfoutput=1

\documentclass[11pt]{article}

\usepackage{acl}

\usepackage{times}
\usepackage{latexsym}

\usepackage[T1]{fontenc}

\usepackage[utf8]{inputenc}

\usepackage{microtype}

\usepackage{inconsolata}

\usepackage{amsmath}
\usepackage{graphicx}
\usepackage{booktabs}
\usepackage{cleveref}
\usepackage{color, colortbl}
\usepackage{multirow}
\usepackage{algorithm}
\usepackage[noend]{algpseudocode}

\newcommand{\ourtaskfull}{Motivational Interviewing}
\newcommand{\ourtask}{MI}
\newcommand{\ourmethodfull}{Dialogue Strategy Learning via Inductive Reasoning}
\newcommand{\ourmethod}{DIIR}
\newcommand{\ourmethodfullwithacronym}{DIIR (\textbf{Di}alogue Strategy Learning via \textbf{I}nductive \textbf{R}easoning)}

\title{Few-shot Dialogue Strategy Learning\\for \ourtaskfull{} via Inductive Reasoning}

\def\authorspace{\hspace{4mm}}
\def\organizationspace{\hspace{3mm}}
\def\ucsdmark{$^1$}
\def\allenaimark{$^2$}
\def\sonymark{$^3$}
\author{
    Zhouhang Xie\ucsdmark{}\authorspace{}
    Bodhisattwa Prasad Majumder\allenaimark{}\authorspace{}
    Mengjie Zhao\sonymark{}\authorspace{}
    \\
    \textbf{
    Yoshinori Maeda\sonymark{}\authorspace{}
    Keiichi Yamada\ucsdmark{}\authorspace{}
    Hiromi Wakaki\sonymark{}\authorspace{}
    Julian McAuley\ucsdmark{}\authorspace{}
    }
        \\
        \ucsdmark{}~UC San Diego\organizationspace{}
        \allenaimark{}~Allen Institute for AI\organizationspace{}
        \sonymark{}~Sony Group Corporation\organizationspace{} \\
        \{\href{mailto:zhx022@ucsd.edu}{zhx022}, 
        \href{mailto:jmacauley@ucsd.edu}{jmcauley}\}@ucsd.edu
        }

\begin{document}
\maketitle

\begin{abstract}

We consider the task of building a dialogue system that can motivate users to adopt positive lifestyle changes: \ourtaskfull{}. 
Addressing such a task requires a system that can infer \textit{how} to motivate a user effectively.
We propose \ourmethod{}, a framework that is capable of learning and applying conversation strategies in the form of natural language inductive rules from expert demonstrations.
Automatic and human evaluation on instruction-following large language models show natural language strategy descriptions discovered by \ourmethod{} can improve active listening skills, reduce unsolicited advice, and promote more collaborative and less authoritative responses, outperforming various demonstration utilization methods.

\end{abstract}

\section{Introduction}

\begin{figure}[tb]
\centering
\includegraphics[clip, width=0.90\linewidth]{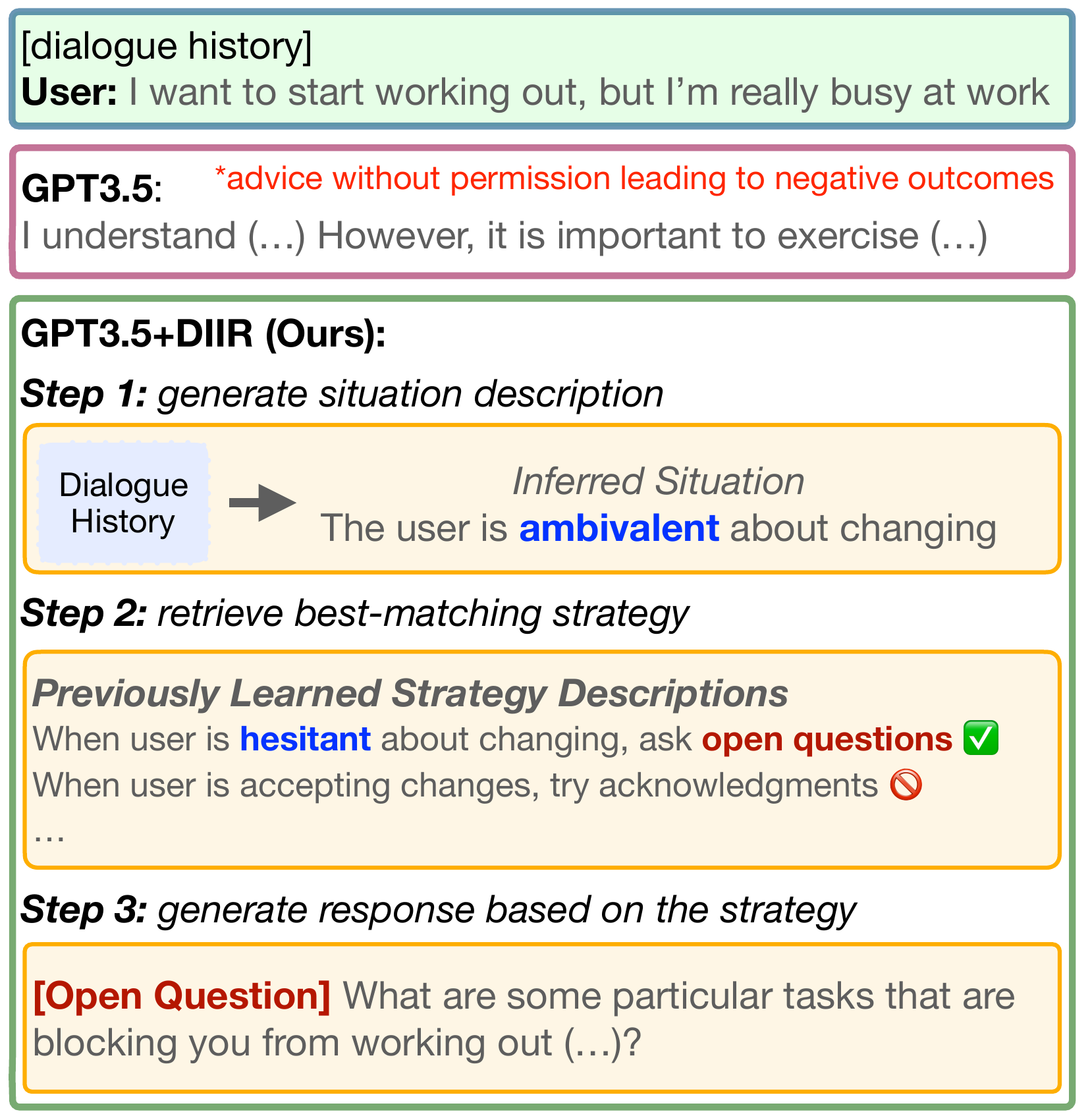}
\caption{\small \ourmethod{} retrieves learned dialogue strategy descriptions to generate a response that \textbf{leads to positive outcomes}.
See \Cref{tab:learned_strategy_descriptions} in the Appendix for example strategies.
}
\label{fig:motivating_situation}
\vskip -5mm
\end{figure}

We study \ourtaskfull{}~(\ourtask{}), where a conversational agent assists users in finding motivation and reducing ambivalence in adopting positive lifestyle changes such as quitting drug abuse and seeking medical assistance~\cite{Moyers2016TheMI}.
A successful MI dialogue system must be able to infer \textit{how} to best motivate a user to adopt positive changes, which is a nontrivial task.
For example, as shown in~\Cref{fig:motivating_situation}, one way to convince a user to workout is to educate the user about its benefits, yet immediately offering suggestions without understanding the context can risk creating resistance and confronting users~\cite{Brehm1981ATO}.
While prior work studied rule-based dialogue systems for MI~\cite{welch-etal-2020-expressive, Park2019DesigningAC} and MI dialogue response rephrasing~\cite{welivita-pu-2023-boosting}, another important problem is to build dialogue systems that \textit{automatically discover and learn} \ourtask{} skills.

In this work, we introduce~\ourmethodfullwithacronym{}, a framework for aligning LLMs to dialogue behavior of professional MI interviewers given a few demonstration dialogues.
\ourmethod{}~does not require training and instead employs LLMs to generate a set of natural language dialogue strategy descriptions, such as \texttt{``when the user is hesitant about change, the interviewer should ask open questions to learn more about the user's concern"} by analyzing demonstration dialogues.
These strategies are hypothesis about how professional interviewers respond in various situations, and are reused at inference time as instructions for LLMs to mimic  MI interviewers.
This process of identifying underlying principles from observations and generalizing them for inference is a form of inductive reasoning~\cite{Feeney2007InductiveRE}.

Different from common practice that use available demonstrations as in-context-learning (ICL) examples~\cite{brown2020lmarefewshot, xu2023knn}, strategies descriptions discovered by our framework explicitly state the desired behavior for the model, such as \texttt{``ask open questions"} and their applicable situations, such as \texttt{``when the client is hesitant about change"}.
These automatically generated strategy descriptions are a more precise nudge to the LLM being instructed than ICL examples.
Existing work that employs language models for generating and reusing natural language strategies from observations~\cite{majumder2023clin, DBLP:conf/icml/NottinghamAS0H023, Wang2024HypothesisSI} (i.e. Inductive Reasoning with LLMs) require interactive environments to verify and improve the generated statements.
In comparison, our method learns from a static demonstration dialogue dataset which is more commonly available than interactive environments for dialogue models.

To evaluate \ourmethod{}, we automate and adapt a set of metrics derived from clinical psychology literature for evaluating human \ourtask{} practitioners.
We experiment with \ourmethod{} on a publicly available \ourtask{} dataset.
Automatic and human evaluation show strategies discovered by \ourmethod{} can help instruction-following downstream language models produce more collaborative, respectful, and least resistance-inducing responses with as few as five annotated dialogues of expert demonstrations.
Our \textbf{contributions} are as follows: 
(1) To our best knowledge, we are the first to work to build and evaluate a dialogue system that automatically learns from dialogue data for \ourtask{}.
(2) We develop \ourmethod{}, a framework that learns and applies dialogue strategies as natural language inductive rules from \ourtask{} experts for only a few dialogues.
(3) We automate evaluations from the clinical psychology literature on \ourtask{} to facilitate future research on \ourtask{}.

\section{\ourmethodfull{}}
\label{sec:ourmethod}

The goal of an MI interviewer is to convince a client to make a positive change of interest, such as to start working out.
Given a dialogue history $h$ consisting of previous user-system turns, an ideal MI dialogue system produces the next system response $a$ that best convinces the user to make a positive change. 
We solve this optimal response generation problem by using LLMs to analyze demonstration dialogues and generate dialogue strategy descriptions stating how to respond in various situations, such as  ``\texttt{When the user is hesitant about change, ask open questions.}'' 
These descriptions can then be reused to improve an instruction-following LLM to promote demonstration-aligned dialogue behavior, such as asking open questions when the client is hesitant.
We now describe \ourmethod{}'s learning and inference process (see  \Cref{sec:step_by_step_learning_and_inference} for step-by-step concrete examples).

\begin{algorithm}[t!]
\small
\caption{\textsc{EnhanceStrategy}}
\label{alg:enhance_strategy}
\begin{algorithmic}[1]
    \Require Dialogue history $h$, gold response $a^*$, generator LLM $G$, discriminator LLM $D$, executor LLM $E$, maximum number of trials $N$
    \Ensure Strategy $e$ in natural language, description of its applicable situation $k$
    \State $e \gets$ ""
    \For{i in $N$}
        \State $a' \gets$ E.\texttt{generate\_response} (h, e)
        \If{D.\texttt{confirms\_is\_similar} ($a'$, $a^*$)}
            \State \textbf{break}
        \EndIf
        \State $e \gets$ G.\texttt{improve\_strategy} ($h$, $e$, $a^*$, $a'$)
        
    \EndFor
    \State $k \gets$ G.\texttt{describe\_situation} ($h$)
    \State \textbf{return} $e, k$
\end{algorithmic}
\end{algorithm}

\noindent
\textbf{Learning Dialogue Strategy by Analyzing Demonstration Dialogues} \hspace{0.7em} 
Given a dialogue history $h$ and a gold interviewer response $a^*$, a corresponding dialogue strategy description $e$ should contain sufficient information such that an LLM could closely reproduce $a^*$ given $h$ and $e$.
Intuitively, one could instruct an LLM to analyze the gold response and its corresponding dialogue history to generate descriptions containing information about ``\texttt{When <description about $h$>, one should <description about $a^*$>},'' which can then be used as strategy descriptions.
However, there is no guarantee that LLM-generated strategy descriptions are valid.

To address this issue, we propose to employ a \texttt{discriminator} LLM $D$ and an \texttt{executor} LLM $E$  in place of the interactive environment in previous works to give feedback during the strategy statement generation process.
The intuition is that to verify a dialogue strategy such as \texttt{``When the user is hesitant about change, ask open questions"}, we could instruct an \texttt{generator} LLM $G$ to generate a response given the dialogue history where a user is hesitant and this instruction, and check whether the generated response matches the gold interviewer response.

This verification process eliminates two sources of error: the generated strategy is incorrect or is correct but is not detailed enough to instruct an LLM to reproduce the groundtruth response.
Existing work typically addresses this statement-verification problem by verifying the statements in interactive environments such as MineCraft~\cite{DBLP:conf/icml/NottinghamAS0H023}. 
However, there are no environments to verify statements about dialogue strategies.

The process for generating and verifying a strategy description given a context-and-gold-response pair is as shown in~\Cref{alg:enhance_strategy}.
Concretely, given a hypothesized dialogue strategy $e$ (initialized to be an empty string), the \texttt{executor} generates a response $a'$ given the dialogue history $h$ and the strategy $e$. 
If the \texttt{discriminator} checks and marks $a'$ as similar to gold response $a^*$, we accept the strategy $e$ as valid. 
Otherwise, we instruct the \texttt{generator} LLM to regenerate a better strategy given the dialogue history, current strategy, generated response, and the gold response.
This generate-and-verify process repeats until a valid strategy description has been found or a fixed number of search steps $N$ has been reached.

Finally, to be able to reuse this strategy in the future, we prompt the \texttt{generator} to generate a description of the situation $k$ such as ``\texttt{the user is hesitant about change}.'' 
This situation description can be derived from $e$ since the dialogue strategy already contains information about when the strategy is applicable, but we use an LLM to generate it in practice for ease of implementation.
We repeat the process introduced above for all context-response pairs available in the demonstration dialogues to create suitable strategies for various scenarios.
While drawing inspirations from adversarial training~\cite{DBLP:conf/nips/GoodfellowPMXWOCB14} and reference games~\cite{fried-etal-2018-unified}, our algorithm is purely language-based.

\paragraph{Inference via Strategy Reuse}

After learning, \ourmethod{}~accumulates a set of strategies $\{e_i\}$
and their corresponding applicable situations
$\{k_i\}$. 
Thus, given an inference-time dialogue history $h_{\text{inference}}$, we can simply generate a description of the situation $k_{\text{inference}}$ using the \texttt{generator} LLM, and then use $k_{\text{inference}}$ to retrieve the most suitable strategy $e_i$ by identifying $k_i$ that is most similar to $k_{\text{inference}}$. 
We then instruct the \texttt{executor} to generate a response given $h_{\text{inference}}$ and the retrieved $e_{i}$.
In practice, we use a text embedding model $\Phi$ to encode all situational descriptions and find the top ten most suitable demonstration situation by maximizing the dot product of $\Phi(k_i)$ and $\Phi(k_\text{inference})$ following
\citet{DBLP:conf/nips/LewisPPPKGKLYR020, Guu2020Realm}.
After this, we instruct the \texttt{Generator} LLM to pick a strategy that is most appropriate to the dialogue history from the ten retrieved strategies and follow its description.
This process is illustrated in~\Cref{fig:motivating_situation}.

\section{Experiments}

\paragraph{Models and Baselines} 
We demonstrate \ourmethod{}'s few-shot learning ability by learning from 5 dialogues using GPT3.5 and GPT4.
We chose these two models due to their strong reasoning ability, similar to various recent works on LLM-based reasoning~\cite{wang2023promptagent, majumder2023clin, DBLP:conf/icml/NottinghamAS0H023}.
For baselines, we consider three variants of in-context learning (ICL) in addition to the vanillan LLM: ICL-RAND with 5 random demonstrations, ICL-KNN~\cite{xu2023knn} with 5 demonstrations with most similar dialogue context to the inference-time context computed using a text embedding model, and ICL-ALL with all demonstration dialogues.

\paragraph{Dataset} 
We experiment on the AnnoMI dataset~\cite{Wu2022AnnoMI}, an MI dataset transcribed from YouTube videos with 133 MI transcripts for various topics labeled as high and low-quality interviewer demonstrations.
We discard all dialogues labeled as low-quality demonstrations, randomly sample 80 dialogues with 1192 turns for our main evaluation on GPT3.5, and use the rest for learning.
For all high-cost experiments involving GPT4 or human evaluation, we further randomly sub-sample 10 dialogues with 148 turns.

\section{Evaluation}

For \textbf{automatic evaluation}, we evaluate the alignment of our model by obtaining dialogue action counts with a dialogue act classifier trained on crowd-sourced annotations~\cite{welivita-pu-2022-curating}.
We report the following behavior-count-based evaluation adopted from the Motivational Interviewing Integrity Treatment (MITI; \citet{Moyers2016TheMI})
and Motivational Interviewing Skill 
Code (MISC; \citet{miller2003manual})
used to evaluate professional interviewers from clinical psychology literature.

\noindent
\textbf{Ratio of MI-guideline-inconsist Behavior (\%MI-i)}\hspace{0.7em}
The ratio of MI-inconsistent behaviors (confront, direct, warn, and advice without permission) in model responses.

\noindent
\textbf{Complex over Simple Reflections (C/S)}\hspace{0.7em}
The ratio of complex over simple reflections in model responses, indicating how often the model produces in-depth feedback to the client. 

\noindent
\textbf{Reflection over Questions (R/Q)}\hspace{0.7em}
The count of reflections over questions in the models' responses.

\noindent
\textbf{Percent of Active Listening Behaviors (\%AL)}\hspace{0.7em}
The ratio of the sum of Questions and Reflections among all behavior counts, showing how often does the modal employ active listening skills.

\noindent
\textbf{Percent of Non-authoritatitive Behaviors (\%NA)}\hspace{0.7em}
The frequency of responses that are not confront, warn, direct, advice (with or without permission), or give information.

We verify the effectiveness of our automatic evaluation by running the evaluation on high and low-quality interviewer demonstrations in the AnnoMI dataset~\cite{Wu2022AnnoMI}, results show our method can effectively tell apart high versus low quality demonstrations (see~\Cref{sec:appenxix_additiona_evaluation_details} for additional discussions and details of the dialogue act classifier). 

For \textbf{human evaluation}, we evaluate our models' performance against in-context baselines by asking human evaluators on Amazon Mechanical Turk to read four context-response pairs produced by both models and choose the one that is best aligned with a description of MI (details in \Cref{sec:appenxix_additiona_evaluation_details}).

\section{Results and Analysis}

\begin{table}[tb]
\small
\setlength\tabcolsep{2.0pt}
\center
\begin{tabular}{lccccc}
\toprule
\bf Method &  \bf \%MI-i↓ &  \bf C/S↑ & \bf R/Q↑ & \bf \%AL↑ & \bf \%NA↑  \\
\midrule
\rowcolor{lightgray}\multicolumn{6}{l}{\textbf{\ W/o Strategies}} \\
GPT3.5-BESTBASE  &  2.1 & 44.8 &  \textbf{1.53} & 54.3 & 82.8  \\
GPT4-BESTBASE  &  1.4   &  11.8 &   1.06 & 45.8 & 71.8  \\
\midrule
\rowcolor{lightgray}\multicolumn{6}{l}{\textbf{\ With Strategies (Ours)}} \\
GPT3.5-DIIR & \underline{1.3} & \underline{115} & 0.99 & \textbf{61.1} & \textbf{95.1}\\
GPT4-DIIR & \textbf{0.6} & \textbf{135} & 1.04 & \underline{59.6} & \underline{94.4} \\
\midrule
Gold &  1.4 & 2.3 &  1.51 & 59.0 & 81.7 \\
\bottomrule
\end{tabular}
\caption{\small Alignment performance against best possible result \textit{selected per dimension} from all baselines. We found open-source LLMs cannot consistently follow self-generated statements, coherent with recent findings~\cite{qiu2024phenomenal}, and thus are less performant (See \Cref{sec:open_source_model}).}
\label{tab:alignment_performance}
\vskip -4mm
\end{table}

Our main results are as shown in~\Cref{tab:alignment_performance}.
Due to space constraints, we report the best performance among the four baselines (base model and three ICL variants) for each dimension under the name BESTBASE, and show detailed performance in~\Cref{tab:alignment_performance_full} in the Appendix.
Although BESTBASE has an unfair advantage due to combining best baseline results for each dimension, our method outperforms the baselines in all categories except the Reflection-to-Question ratio, sometimes outperforming the ground truth responses.
Meanwhile,
MI guideline defines an R/Q ratio of 1.0 to be good and 0.5 to be satisfactory~\cite{Moyers2016TheMI}, showing that our model maintains competitive performance in this dimension.

Overall, our method produces fewer MI-inconsistent responses, provides more in-depth reflections, and better leverages active listening skills and the MI spirit of not assuming expert roles.
These trends are coherent with human evaluation results, as shown in~\Cref{tab:human-eval}.

\begin{table}[t!]
\small
\center
\begin{tabular}{lcccc}
\toprule
{DIIR vs. ICL}  & \bf  RAND & \bf ALL & \bf KNN  \\
\midrule
GPT3.5  & \textbf{21}* & \textbf{7}*  & \textbf{23}* \\
GPT4  &  \textbf{15}* & \textbf{5} & -5 \\
\bottomrule
\end{tabular}
\caption{\small Human evaluation results. We report number of \textit{additional} times \ourmethod{}~win against baseline; with statistical difference with p$<$0.05 denoted in *.
}
\vskip -5mm
\label{tab:human-eval}
\end{table}

\section{Related Work}

\paragraph{Motivational Interviewing in NLP} 
Prior work in NLP about MI primarily studied the task of annotating~\cite{welivita-pu-2022-curating, tanana-etal-2015-recursive} and predicting~\cite{perez-rosas-etal-2017-predicting, huang-etal-2018-modeling} dialogue actions.  
Recently, there are attempts on building systems for evaluating~\cite{min-etal-2022-pair} and rewriting~\cite{welivita-pu-2023-boosting} responses for MI but does not focus on response generation.
A few pioneering works build MI dialogue systems with hard-coded rules or dialogue templates~\cite{welch-etal-2020-expressive, Park2019DesigningAC} but focuses specific domains such as COVID-19.
To the best of our knowledge, we are the first to build an MI dialogue system that automatically learns from expert demonstrations.

\paragraph{LLMs for Generating and Reusing Natural Language Strategies}
Generating hypothesis from observations and reusing the generated hypothesis with LLMs have shown a promising direction for various tasks, such as planning~\cite{majumder2023clin,DBLP:conf/icml/NottinghamAS0H023, xiang2023language}, prompt-generation~\cite{wang2023promptagent}, and code generation~\cite{Wang2024HypothesisSI}.
This is a form of inductive reasoning~\cite{xiang2023language} where LLMs draw conclusion from obvervations.
However, prior works require interactive environments for feedback which are unsuitable for learning from static datasets, while our work use an LLM for feedback and lifts the requirement for interactive environments.

\section{Conclusion}

We propose \ourmethod{}, a method for few-shot aligning large language models to \ourtask{} techniques.
Our method infers natural language strategies from expert demonstrations and reuses these strategies at inference time.
Automatic and human evaluation show our method is better aligned with the behavior of motivational interviewing experts, produces more in-depth reflections, and better leverages active listening skills, outperforming a variety of demonstration-based methods.

\section{Limitations}

We note that our framework is developed for \ourtaskfull{} and benefits from domain-specific knowledge such as \ourtask{}-specific dialogue actions.
Meanwhile, our framework assumes access to a strong instruction following language model, which is usually restricted by API access or computational resources.
Finally, we note that while we adopt clinical psychology literature verified metrics and human evaluation to measure the performance of DIIR, the ideal evaluation is deploying MI dialogue systems for continued intervention.

\section{Ethical Concerns}

Following previous work on conversational persuasion~\cite{wang-etal-2019-persuasion}, we note \ourtaskfull{} shares similarities with persuasion dialogues.
While the technique for \ourtaskfull{} is developed for inducing \textit{positive} changes in users, system deployers should carefully analyze the specific topic for \ourtaskfull{} to eliminate risks of harm.
Meanwhile, when deploying a system equipped with \ourtask{} abilities, the user should be informed about the system's identity so that they can make an informed choice when interacting with such a system.

\section{Acknowledgments}

We would like to thank Tim Althoff, Jessica Echterhoff, and Yilan Chen for their helpful comments. This work is funded by the SONY Research Award Program.

\bibliography{myanthology,custom, references}

\appendix

\section{Building the Dialogue Act Classifier}
\label{sec:dialogue_act_classifier}
Following prior work in the Computer Science and NLP community on annotating motivational interviewing dialogues~\cite{Shah2022ModelingMI, welivita-pu-2022-curating, perez-rosas-etal-2016-building, tavabi-etal-2021-analysis}, we train a dialogue classifier to map sentences to motivational interviewing behavior codes.
We train the dialogue act classifier on a publicly available dataset~\cite{welivita-pu-2022-curating} with annotated dialogue actions specific to \ourtask{}, since the AnnoMI dataset does not contain fine-graned dialogue action labels.
This dataset contains single turn interactions only, and thus are suitable for training dialogue act classifiers only but not the dialogue model itself.
We fine-tune two variants of pre-trained mask language models, BERT~\cite{Vaswani2017AttentionIA} and Mental-BERT~\cite{ji-etal-2022-mentalbert}, a fine-tuned version of the base BERT model on psychotherapy dialogues.
We use a $0.6-0.2-0.2$ train-validation-test split.
We fine-tune both models using the Huggingface Transformers Library~\cite{wolf-etal-2020-transformers} for a maximum of 10 epoch with early stopping at a learning rate of $5e-5$.
While the absolute accuracy of both BERT~\footnote{https://huggingface.co/bert-base-uncased} and Mental-BERT~\footnote{https://huggingface.co/mental/mental-bert-base-uncased} are not optimal (Recall at 1 at 0.60 and 0.61, respectively), we fine the Recall at 5 out of 12 classes are high (at 0.95).
Thus, we use the classifier to retrieve N=5 relevant labels given a input sentence, and use a GPT-3.5-turbo to decide the final label, following recent findings in NLP that combining an LLM with small classifier for classification has superior performance to both base models~\cite{xu2023small}.

\section{Other Experiment and Implementation Details}
\label{sec:implementation_details}
For \ourmethod{}'s learning process, we set the maximum number of optimization steps to 3, and break the reference game between the \texttt{generator} and the \texttt{discriminator} afterwards regardless of success.
To enhance the feedback quality of the \texttt{discriminator}, we also use the same dialogue act classifier as described in~\Cref{sec:dialogue_act_classifier} to provide distant labels of what potential actions are in the ground truth response (this means predicting the string labels of dialogue actions, and append this to the instruction for the LLM).

For all experiments, the specific model names we used in OpenAI is \texttt{gpt-3.5-turbo-1106} and \texttt{gpt-4-1106-preview}.
These models are available as of December 2023.
For the off-the-shelf retriever for retrieving strategies, we use a publicly available light-weight model~\footnote{https://huggingface.co/sentence-transformers/all-MiniLM-L6-v2} and retrieve usiang dot product scores.
All experiments are conducting using 1  RTX A6000 GPU. 
Learning and Inference of \ourmethod{} on the specified data size finishes in 3 hours and 1.5 hours, respectively.
The trainable component of the dialogue act classifier (based on BERT) contains 12-layer, 768-hidden, 12-heads, and 110M parameters.
The retriever model contains 12-layer, 384-hidden, 12-heads, 33M parameters.

\section{Details on Evaluation}
\label{sec:appenxix_additiona_evaluation_details}

\subsection{Discussion on Evaluation Metrics}

We provide discussions on the derived metrics in this section.
In particular, the evaluation process in MISC evaluates a interviewer on three dimensions: (1) Acceptance: being non-judgemental and do not direct or confront the user, (2) Empathy (different from the definition as used in Empathetic Dialogues in NLP): showing genuine interest in understanding the client's situation and thoughts with active listening skills, and (3) Motivational Interviewing Spirit: the interviewer should be collaborative rather than assuming expert roles.
In practice, we found all models rarely direct or confront the user using our dialogue action classifier, trivially satisfying the Acceptance criteria.
We use \%AL as a proxy for the Empathy criteria, and \%NA as a proxy to measure the Motivational Interviewing Spirit criteria.
These metrics are proven to reduce client resistance and improve intervention outcome in MI literature~\cite{Moyers2016TheMI}.

\subsection{Verifying the Effectiveness of Automatic Evaluation}

\begin{table}[tb]
\small
\setlength\tabcolsep{2.0pt}
\center
\begin{tabular}{lccccc}
\toprule
\bf {} &  \bf \%MI-i↓ &  \bf C/S↑ & \bf R/Q↑ & \bf \%AL↑ & \bf \%NA↑  \\
\midrule
Low  &  20 & 2.09 &  1.0 & 34.4 & 51.5  \\
High  &  \textbf{0.9}   &  \textbf{2.75} &  \textbf{1.875} & \textbf{62.4} & \textbf{82.6}  \\
\bottomrule
\end{tabular}
\caption{Our automatic evaluation can tell apart between high quality (``High" in the table) and low quality (``Low" in the table) MI demonstrations}
\label{tab:evaluating_eval_method}
\vskip -2mm
\end{table}

to validate our evaluation, we run our classifier-based evaluation pipeline on 231 datapoints (dialogue context + gold response) of good vs. bad demonstration of MI interviewers (the binary high v.s. low quality labels are available in the original AnnoMI dataset). Bad demonstrations here means the response are produced by interviewer that does not follow MI guideline. We use 231 pairs from each class (i.e. 231 high and 231 low-quality demonstrations) since that is the number of all bad demonstrations available, and we downsampled good demonstrations to balance classes. Specifically, the desired behavior here is our evaluation pipeline should be able to tell apart good versus bad MI interviewers.

As shown in \Cref{tab:evaluating_eval_method}, our evaluation pipeline does indicate that high quality MI interviews are much better than low quality ones - this shows that our evaluation method can tell apart good vs bad MI interviewers.

\subsection{Human Evaluation Details}

We use Amazon Mechanical Turk for our human evaluation\footnote{https://www.mturk.com/}.
The specific prompt we use is \texttt{``Which response does not claim an expert role and make overt and frequent attempt to persuade personB, doesn't push for persuation when giving factual information, and act more like a companion that show genuin interest in understanding personB's thoughts and situation? Pick the one that you thinks is best described by the style above"}.
To unsure quality response, we request works to have an lifetime approval rate of 98\% or above, and filter out all response that failed an attention check (for example, picking an animal that is not often kept as pet from ragdoll cat, golden retriever, elephant, and gold poodle). 
We do not impose any constraints (other than English-speaking), and pay at least 0.1 dollar per task (8 responses).

\section{Concrete Example of \ourmethod{}'s Learning and Inference Process}
\label{sec:step_by_step_learning_and_inference}

\subsection{Learning}

We provide a concrete example of the learning process in the “Inferring Strategy” section in \Cref{tab:learning_with-gpt-3.5}.
The comments denoted in \texttt{``\#\#"} in the table should make it self-contained, but we provide more details here.
The input to our model is the dialogue state (dialogue history $h$). First, the \texttt{generator} (GPT-3.5-Turbo) generates an Initial Response \texttt{``It is normal to…”} prompted with the dialogue context; this is just a vanilla response generated by the LLM. Then, the \texttt{discriminator} model (another LLM) checks and gives feedback to the generator LLM on its flaws compared to the ground truth, in this case \texttt{``when the client acknowledges the difficulty…”}. The generator will then be prompted to regenerate the response based on Dialogue State as well as the feedback from the \texttt{discriminator} model. In this case, the regenerated response is \texttt{``It sounds like it’s been…”} At this point, the \texttt{discriminator} model will be prompted to check whether the response is matching ground truth - here, both the generated response \texttt{``It sounds like it’s been…”} and the gold response \texttt{``It’s something that’s always on your mind”} are reflecting on what the client said, and thus the \texttt{discriminator} marked the process as complete. We then save this instruction \texttt{``when the client acknowledges the difficulty…”} as a stategy decription $e$ for reuse at inference time. The \texttt{discriminator} might not always generate ideal feedback on the first trial, so we prompt the \texttt{discriminator} model to refine the feedback and repeatedly try the new feedback prompt on the generator until success (or some predefined maximum number of refinement steps reached).

\subsection{Inference}

We provide a concrete example of inference in \Cref{tab:inference_with-gpt-3.5}.
The comments denoted as \texttt{``\#\#"} should make the table self contained, but we provide more details here.
The input to the model is the Dialogue State (dialogue history $h$). %
After this, we prompt an LLM to generate a dialogue situation description of the dialogue state, for example, \texttt{``The client is hesitant and unsure about changing yet"}. 
Recall that at training time we assign a dialogue situation description like the one above to each training dialogue state, so at inference time we can use a dense retriever to pick out the training data that has the most similar client mental states to the current one. We then prompt a separate instance of LLM with the current dialogue state (context) and the N=10 natural language strategies associated with the retrieved dialogue contexts, and let it determine the most relevant one, i.e. the reranking step. In the example shown in Table 5, the selected strategy is \texttt{``when the client seems hesitant and uncertain about making a positive change…"} We give this strategy to the generator model and let it improve the response, producing the final output \texttt{``It’s important to consider the potential risks…"}.

\begin{table}[tb]
\small
\setlength\tabcolsep{2.0pt}
\center
\begin{tabular}{lccccc}
\toprule
\bf Method &  \bf \%MI-i↓ &  \bf C/S↑ & \bf R/Q↑ & \bf \%AL↑ & \bf \%NA↑  \\
\midrule
\rowcolor{lightgray}\multicolumn{6}{l}{\textbf{\ W/o Strategies}} \\
GPT3.5  &  3.1 & 44.8 &  \underline{1.29} & 43.0 & 70.6  \\
GPT3.5-ICL-ALL  &  3.4   &  6.25 &   \textbf{1.53} & 28.4 & 70.6  \\
GPT3.5-ICL-RAND  &  2.8   &  7.05 &   0.98 & 51.7 & 82.3  \\
GPT3.5-ICL-KNN  &  2.1   &  4.28 &   0.93 & 54.3 & 82.8  \\
GPT4  & 1.4 & 11.8 &  0.86 & 41.1 & 71.3  \\
GPT4-ICL-ALL & 2.4 & 10.8 & 0.91 & 38.0 & 64.8 \\
GPT4-ICL-RAND  &  3.0   &  10.55 &   1.01 & 45.1 & 71.8  \\
GPT4-ICL-KNN  &  3.1   &  4.18 &   1.06 & 45.8 & 71.8  \\
\midrule
\rowcolor{lightgray}\multicolumn{6}{l}{\textbf{\ With Strategies (Ours)}} \\
GPT3.5-DIIR & \underline{1.3} & \underline{115} & 0.99 & \textbf{61.1} & \textbf{95.1}\\
GPT4-DIIR & \textbf{0.6} & \textbf{135} & 1.04 & \underline{59.6} & \underline{94.4} \\
\midrule
Gold &  1.4 & 2.3 &  1.51 & 59.0 & 81.7 \\
\bottomrule
\end{tabular}
\caption{Alignment performance (full table). ↑ and ↓ marks denote positive and negative metrics.}
\label{tab:alignment_performance_full}
\vskip -2mm
\end{table}

\section{Discussion on Open Source Model}
\label{sec:open_source_model}

\begin{table}[tb]
\small
\setlength\tabcolsep{2.0pt}
\center
\begin{tabular}{lccccc}
\toprule
\bf Method &  \bf \%MI-i↓ &  \bf C/S↑ & \bf R/Q↑ & \bf \%AL↑ & \bf \%NA↑  \\
\midrule
Mistral  &  5.4 & \underline{22.5} &  1.15 & 44.4 & 67.7  \\
Mistral-RAND  &  5.4   &  2.1 &   \textbf{1.57} & \textbf{45.5} & \textbf{74.5}  \\
Mistral-KNN  &  5.5   &  1.97 &   \underline{1.55} & \underline{45.0} & \underline{73.8}  \\
Mistral-DIIR  &  \textbf{4.8}   &  \textbf{25.2} &   1.14 & \underline{45.0} & 68.4  \\
\bottomrule
\end{tabular}
\caption{We additionally report alignment performance of Mistral, and open source LLM. ↑ and ↓ marks denote positive and negative metrics.}
\label{tab:alignment_performance_mistral}
\vskip -2mm
\end{table}

In our exploration phase, we test \ourmethod{}~on open source models, but find they cannot consistently follow the generated strategy statements.
This is coherent with recent findings that LLMs are better at generation than understanding~\cite{qiu2024phenomenal, west2024generative}.
We provide the performance of Mistral\footnote{https://huggingface.co/mistralai/Mistral-7B-v0.1} in \Cref{tab:alignment_performance_mistral}, as shown in results, our method is comparable to various baselines on a weaker-instruction following model (than GPT3.5 and GPT4).

\section{Definition of Dialogue Actions}

We show definition of each dialogue action as in~\Cref{tab:dialogue-action-definition}.

\begin{table*}[t!]
\small
\center
\begin{tabular}{llcc}
\toprule
\bf Dialogue Action &  \bf Definition \\
\midrule
Give Information  &  \begin{minipage}[t]{1.4\columnwidth}Gives information, educates, provides feedback, or expresses a professional opinion without persuading, advising, or warning. Self-discose of objective information also goes here.\end{minipage} \\ \midrule
Question  &  \begin{minipage}[t]{1.4\columnwidth}All questions from clinicians (open, closed, evocative,	fact-finding, etc.)	\end{minipage} \\ \midrule
Simple Reflection  &  \begin{minipage}[t]{1.4\columnwidth}Reflect (repeat or reword) on what the client have said, without adding further meaning to it.	\end{minipage} \\ \midrule
Complex Reflection  &  \begin{minipage}[t]{1.4\columnwidth}Reflect (repeat or reword) on what the client have said, but adding further meaning (or make explicit some hidden impliciation) of it.	\end{minipage} \\ \midrule
Affirm  &  \begin{minipage}[t]{1.4\columnwidth}States something positive or complimentary about the client’s strengths, efforts, intentions, or worth.\end{minipage} \\ \midrule
Emphasize Autonomy  &  \begin{minipage}[t]{1.4\columnwidth}Highlights a client’s sense of control, freedom of choice, personal autonomy, ability, and obligation about change.\end{minipage} \\ \midrule
Confront  &  \begin{minipage}[t]{1.4\columnwidth}Directly and unambiguously disagreeing, arguing, correcting, shaming, blaming, criticizing, labeling, warning, moralizing, ridiculing, or questioning a client’s honesty.\end{minipage} \\ \midrule
Seek Collaboration  &  \begin{minipage}[t]{1.4\columnwidth}Attempts to share power or acknowledge the expertise of a client.\end{minipage} \\ \midrule
Support  &  \begin{minipage}[t]{1.4\columnwidth}These are generally sympathetic, compassionate, or understanding comments, with the quality of siding with the client.\end{minipage} \\ \midrule
Advise with Permission  &  \begin{minipage}[t]{1.4\columnwidth}Attempts to change a client’s opinions, attitudes, or behaviors, but have obtained the client's permission to do so, or clearly indicates the decision is the clients'.\end{minipage} \\ \midrule
Advise without Permission  &  \begin{minipage}[t]{1.4\columnwidth}Attempts to change a client’s opinions, attitudes, or behaviors using tools such as logic, compelling arguments, self-disclosure, facts, biased information, advice, suggestions, tips, opinions, or solutions to problems.\end{minipage} \\ \midrule
Other  &  \begin{minipage}[t]{1.4\columnwidth}Filler words, such as 'mm-hmm', 'mm', 'yeah', 'okay', 'hmm', 'uh-huh', 'huh', 'right', 'yep', etc.\end{minipage} \\ \midrule
\bottomrule
\end{tabular}
\caption{Dialogue Actions and Their Definition
}
\label{tab:dialogue-action-definition}
\end{table*}

\begin{table*}[t!]
\small
\center
\begin{tabular}{l}
\toprule
\bf Example of Learning w/ \ourmethod{}+GPT3.5  \\
\midrule
\begin{minipage}[t]{2.0\columnwidth}
\textbf{Dialogue State $h$}\newline
Topic: reducing drug use\newline
[therapist]: You don't want to be here.\newline\newline
[client]: I really want to get off the program.\newline
[therapist]: You seem motivated to follow the rules.\newline
[client]: Mm-hmm. I still crave smoking, even though I've stopped.\newline
\newline
\textbf{Gold Response $a^*$}:It's something that's always on your mind.\newline
\newline\newline
\textbf{Inferring Strategy...}\newline 
\#\# Executor $E$ generates a response $a'$ based on dialogue history only (the strategy $e$ is an empty string at this point): \newline\newline
Executor Response $a'$: It's normal to still have cravings even after stopping. Let's talk about some strategies to help you manage those cravings.\newline 
---\newline
\#\# Discriminator $D$ checks the generated response vs. the gold response. The generated response immediately offers solution, which is way different from gold response.\newline\newline 
Discriminator Checks Success Result: No success.\newline 
---\newline
\#\# Now generator tries to generate a better strategy than the current one (the current one is empty string).\newline\newline
Generator: when the client is acknowledging the difficulty of resisting the craving for the bad habit despite having stopped, the therapist should use a reflective statement to acknowledge the client's struggle and show empathy. The therapist should not jump straight into providing strategies or solutions. The therapist should use 1 sentence for the reflective statement, which should focus on acknowledging the client's struggle and showing empathy.\newline
---\newline
\#\# Now the executor generates a response based on the new strategy description above and the dialogue history.\newline\newline
Executor Response $a'$: It sounds like it's been really tough for you to resist the cravings, even though you've stopped.\newline
---\newline
\#\# Both generated response and the gold response are reflecting on what the client have said now.\newline\newline
Discriminator Check Sucess Result: Yes.
\newline---\newline
\#\# Now we can save the strategy above (``when the client is acknowledging...").\newline
\#\# We further generate a description of the situation for reusing this strategy in the future.\newline\newline 

Situation Description $k$ from Generator: The client is acknowledging the difficulty of change despite making progress.
\end{minipage} \\ 

\bottomrule
\end{tabular}
\caption{Example of \ourmethod{}'s learning process.
}
\label{tab:learning_with-gpt-3.5}
\end{table*}

\begin{table*}[t!]
\small
\center
\begin{tabular}{l}
\toprule
\bf Example of Inference w/ \ourmethod{}+GPT3.5  \\
\midrule
\begin{minipage}[t]{2.0\columnwidth}
\textbf{Dialogue State $h$}\newline
Topic: reducing alcohol consumption\newline
[therapist]: Yeah, you're surprised to hear that?\newline
[client]: Yes. What-what kind of health problems?\newline
[therapist]: Well things like heart disease, cancer, liver problems, uh, stomach pains, insomnia. Unfortunately, uh, people who drink at a risky level are more likely to be diagnosed with depression and alcohol can make depression worse or harder to treat.\newline
[client]: Hmm. Well, that's not good news.
\newline\newline
\textbf{Inferred Client Mental State (Situation Description) $k_{inference}$}:\newline 
\#\# Note we will use the dialogue situation description below to retrieve strategies.
\newline The client is hesitant and unsure about changing yet.\newline\newline
\textbf{Retrieved Strategy $e_i$}:\newline
\#\# The associated situation description $k_i$ for the strategy below would be ``the client seems hesitant and uncertain about making a positive change"\newline
\#\# Note how the current dialogue state is also about a situation where client is unsure about making a change. \newline 
\#\# In this way, if we encode both $k_{inference}$ and $k_i$, they would have large dot product and thus can be retrieved.
\newline
\textbf{Strategy: }when the client seems hesitant and uncertain about making a positive change in their behavior or bad habit, the therapist should advise the client on the potential risks and benefits of their behavior in one sentence. Then, the therapist should ask the client if they would be open to exploring further information or options in one sentence. The therapist should not immediately seek collaboration or suggest a plan in the first response.\newline\newline 
\#\# Now the executor LM can generate the final response given the retrieved strategy.\newline
\textbf{Updated Response}:It's important to consider the potential risks and benefits of your alcohol consumption. Would you be open to exploring further information or options?
\end{minipage} \\ 

\bottomrule
\end{tabular}
\caption{Example of \ourmethod{}'s inference process.
}
\label{tab:inference_with-gpt-3.5}
\end{table*}

\begin{table*}[t!]
\small
\center
\begin{tabular}{l}
\toprule
\bf Example of Learned strategy descriptions w/ \ourmethod{}+GPT3.5  \\
\midrule
\begin{minipage}[t]{2.0\columnwidth}
when the client is open and willing to engage in the conversation, the therapist should first give information about the confidentiality of the conversation in 1 sentence. Then, the therapist should give information about the purpose of the discussion in 1 sentence. Finally, the therapist should ask a question to ensure the client understands and is comfortable with the conversation in 1 sentence. The therapist should not immediately jump into discussing the client's bad habit without setting the context and ensuring the client's understanding and comfort.\newline\newline
when the client downplays the significance of their bad habit, the therapist should ask a specific question to gather more information about the habit and its impact. The therapist should not make general statements about the importance of understanding the habit's role in the client's life without directly addressing the client's behavior. The therapist should ask a specific question to gather more information about the habit and its impact, which is 1 sentence.\newline\newline
when the client expresses enjoyment or positive aspects of their bad habit, the therapist should reflect the client's feelings in one sentence. The therapist should not immediately suggest alternative activities or changes. Instead, the therapist should follow up with a question to explore any negative aspects or ambivalence the client may have about their bad habit in a second sentence. This approach allows the therapist to fully understand the client's perspective before exploring potential changes.
\end{minipage} \\ 

\bottomrule
\end{tabular}
\caption{Three examples of the strategy descriptions learned.
}
\label{tab:learned_strategy_descriptions}
\end{table*}

\begin{table*}[t!]
\small
\center
\begin{tabular}{l}
\toprule
\bf Prompt for Client Mental Stage Inference  \\
\midrule
\begin{minipage}[t]{2.0\columnwidth}
You are a dialogue analyst and your job is to help us understanding motivational interviewing dialogues.  You will be given a dialogue context, and you will help us determine which of the 5 stages of change the client is at: Precontemplation, Contemplation, Preparation, Action, or Maintenance.\newline\newline

1. Precontemplation: At this stage, the individual is not yet considering making a change and may be unaware of the need for change.\newline

2. Contemplation: In this stage, the individual is aware of the need for change and is considering the possibility of making a change in the near future.\newline

3. Preparation: During this stage, the individual is actively preparing to make a change and may be taking small steps toward behavior change.\newline

4. Action: At this stage, the individual has made a specific, observable change in their behavior and is actively working to maintain this change.\newline

5. Maintenance: In the maintenance stage, the individual has successfully made the desired change and is working to prevent relapse and sustain the new behavior over time.\newline

Look at the following dialogue snippet, which of the 5 stage is the client in?\newline

@snippet@\newline

Format you answer in this format: {'prediction': "your answer"}, you do not have to explain anything.
\end{minipage} \\ 
\bottomrule
\end{tabular}
\caption{The prompt we used for client mental stage inference.
}
\label{tab:prompt_for_tom_inference}
\end{table*}

\section{License Statement}

We plan to release the code base upon acceptance of this work under CC BY license.

\end{document}